\documentclass{amia}
\usepackage{graphicx}
\usepackage[labelfont=bf]{caption}
\usepackage[superscript,nomove]{cite}
\usepackage{color}
\usepackage{booktabs}

\begin{document}
\title{Assessing Social Determinants-Related Performance Bias of Machine Learning Models: 
A case of Hyperchloremia Prediction in ICU Population} 
\author{Songzi Liu, M.S.I.S.$^{1}$, Yuan Luo, Ph.D.$^{1}$}
\institutes{
    $^1$Feinberg School of Medicine, Northwestern University, Chicago, IL, USA\\
}
\maketitle

\noindent{\bf Abstract}
\textit{Machine learning in medicine leverages the wealth of healthcare data to extract knowledge, facilitate clinical decision-making, and ultimately improve care delivery. However, ML models trained on datasets that lack demographic diversity could yield suboptimal performance when applied to the underrepresented populations (e.g. ethnic minorities, lower social-economic status), thus perpetuating health disparity. In this study, we evaluated four classifiers built to predict Hyperchloremia—a condition that often results from aggressive fluids administration in the ICU population—and compared their performance in racial, gender, and insurance subgroups. We observed that adding social determinants features in addition to the lab-based ones improved model performance on all patients. The subgroup testing yielded significantly different AUC scores in 40 out of the 44 model-subgroup, suggesting disparities when applying ML models to social determinants subgroups. We urge future researchers to design models that proactively adjust for potential biases and include subgroup reporting in their studies.}

\section*{INTRODUCTION}

Machine learning (ML) is commonly used to identify patterns, extract knowledge, and make predictions from the massive amount of healthcare data in clinical care. While the constant advancements in methodological design almost guarantee more elaborate models that yield improved performance metrics over time, the model performance by itself does not portray a complete picture when it comes to the fair, responsible, and effective deployment of ML in real-world settings. One crucial yet oft-neglected aspect is that ML algorithm could be biased toward certain populations due to numerous factors\cite{ref1a}; examples include the lack of demographic diversity in training data and feature labeling that inherits biases from the human coders.\cite{ref1,ref2} One study proposed three central themes in ensuring distributive justice in machine learning: equal performance, equal outcomes, and equal allocation of resources.\cite{ref1} However, an extensive literature search revealed that researchers seldom report model performance in demographics subgroups, the knowledge of which could help practitioners to determine if an algorithm is sufficiently generalizable to different populations. In the handful of studies that conducted subgroup reporting, results on the presence and severity of machine learning biases were often mixed. One prominent example is that commercial facial detection algorithms trained on lighter-skins subjects exhibited significant performance disparities in classifying different races and genders.\cite{ref3} In another study, the researchers tested the performance of a Convolutional Neural Network model built to predict low left ventricular ejection fraction from the ECG data in racial, gender, and insurance subgroups; the results demonstrate no significant difference in performance metrics.\cite{ref4}

Systematic bias in ML algorithms could perpetuate health disparities.  Models built without adjusting for the structurally biased data may make disproportionately more prediction errors in the under-represented population such as racial minorities. In a study designed to investigate the performance bias in machine learning models trained with unstructured clinical notes, researchers confirmed the difference in prediction accuracy with regard to gender and insurance types for ICU mortality.\cite{ref5} Obermeyer et al. dissected a commonly used algorithm in population health management and identified significant between-race differences: at any given risk score, black patients were shown to be considerably sicker comparing to their white counterparts.\cite{ref6} Bhavani et al. showed that during the COVID-19 pandemic, Black patients in ICU had higher SOFA scores and higher prevalence of comorbidities. If there was a severe scarcity of ventilators, the ventilator allocation protocols currently used by many states would assign lower priority tiers and significantly less allocation of ventilators to Black patients, leading to significantly lower survival.\cite{ref6a}  ML applications with biased performance further compromise the care for the already disadvantaged under-represented demographic groups, broadening the gap in health outcomes between different subgroups. Therefore, assessing and addressing potential biases in ML applications is imperative in ensuring the fairness and effectiveness of ML-assisted clinical practice. 

The administration of intravenous (IV) fluids in critical care setting has long been thought of as a commonplace and safe treatment approach for patients with electrolyte imbalances and fluid deficits. Nevertheless, numerous studies have demonstrated the link between excessive use of IV fluids and less favorable ICU outcomes, including a greater rate of in-hospital mortality, organ damages, and a higher risk of developing acute kidney injury (AKI).\cite{ref7,ref8,ref8a} One particular condition often associated with the aggressive fluid administration approach is hyperchloremia, where an excessive amount of chloride content accumulates in the body as a result of chloride-rich IV infusions. hyperchloremia is shown to be associated with unfavorable clinical outcomes in numerous studies.\cite{ref9,ref10} A relatively new area of investigation, hyperchloremia's impacts were far from being extensively studied. Specifically, there has been a lack of focus on the disease regarding social determinants of health (SDOH) factors such as gender, race, and socioeconomic status (SES). One study explored the racial difference in the conservative fluids management approach's long-term outcomes and observed decreased mortality in non-Hispanic black patients.\cite{ref11} Yet the insufficient sample size and the lack of comparable studies lessened the generalizability of the finding.  Questions such as whether certain racial groups or patients with lower SES profiles were at higher risk of developing hyperchloremia were left unanswered.  Thus, we believe that a more comprehensive understanding of hyperchloremia would have far-reaching implications on fluid management strategy and improved ICU outcomes.

\section*{METHODS}
\subsection*{Study Design}
In our previous studies, we trained predictive four supervised learning classifiers—Ridge Regression, Random Forest, XGBoost, and Multi-layer Perception—for predicting hyperchloremia with features representative of clinical records from the first 24h of adult ICU stays.\cite{ref12} We used retrospective data from MIMIC III, a well-studied dataset containing over 40,000 ICU encounters from 2001 to 2012.This study built on the previously derived classifiers and aimed to validate whether they perform indiscriminately and uniformly on subgroups with distinct demographic traits and whether including SDOH features in addition to clinical indicators set would significantly increase performance across different models. In addition, we looked further into the interpretability of Models by visualizing Feature importance  of the full model and four single-race-trained models using the SHAP method. 
\begin{figure}[]
\centering
\includegraphics[width=\textwidth,height=\textheight,keepaspectratio]{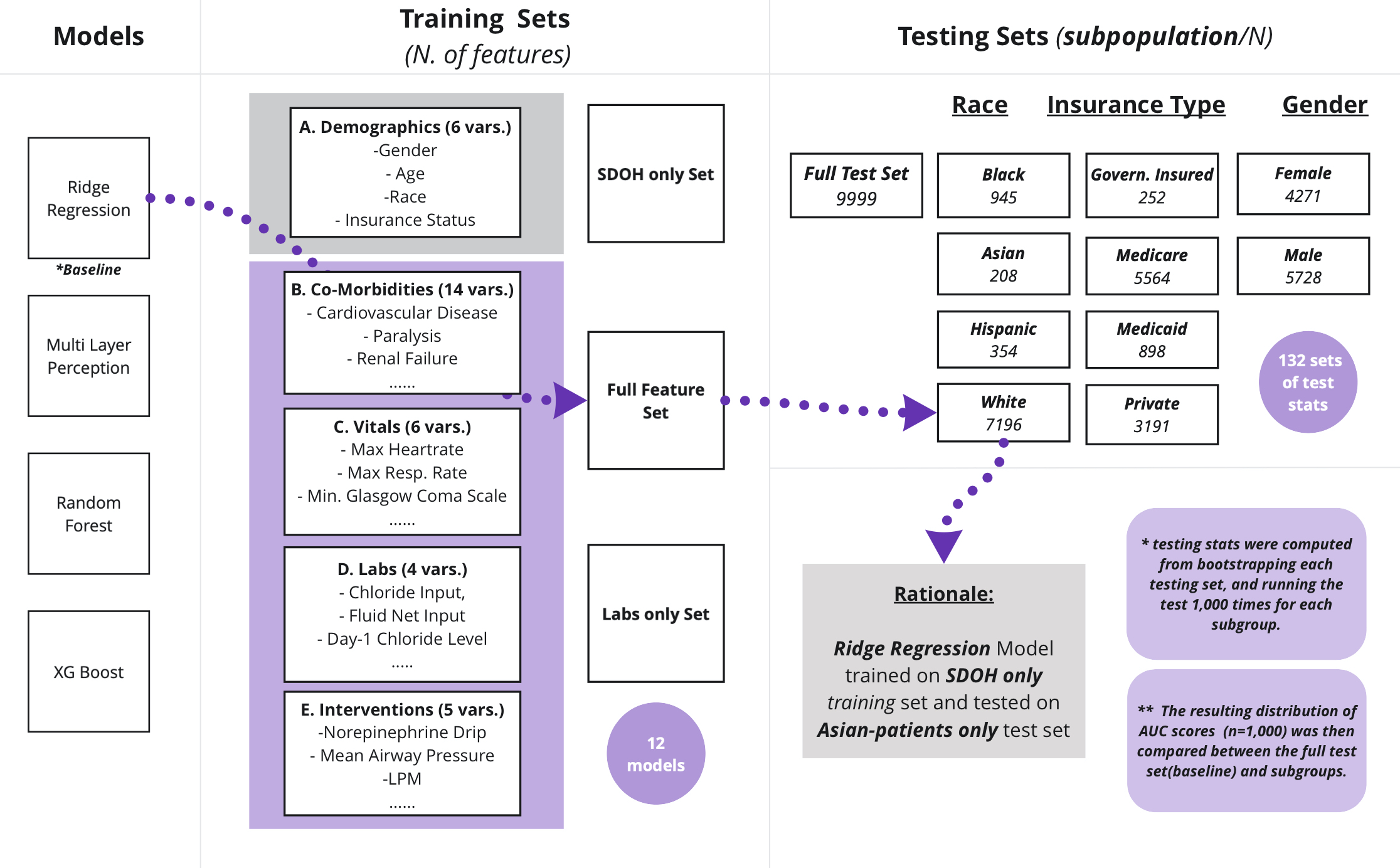}
\caption{Breakdown of Features and Testing Set }
\label{fig1}
\end{figure}
\subsection*{Model Building}
All feature selection criteria, model tuning features, and cut-off values were obtained from the previously published models. We selected 34 features, each with statistically significant differences (p-value < 0.05) between hyperchloremic and non-hyperchloremic patients on day 2, including demographics, co-morbidity upon admission, chloride-related data (net fluid balance, total chloride load, maximum serum chloride), laboratory test results, interventions, medications, and vitals. A patient is labeled hyperchloremic if their record displayed a serum chloride measurement of 110 mEq/L or greater on the second day of their ICU stay. To compensate for the low prevalence of Hyperchloremia (5.98\%, full cohort), we configured the ridge regression and XGBoost classifiers to assign weights based on prevalence. For the random forest and multi-layer perception models, we down-sampled the training set by randomly removing 90\% of non-hyperchloremic patients, resulting in a sufficiently large prevalence of 38.29\%. 

In creating the original study cohort, we excluded patients under the age of 18 and re-admission records given the consideration that such groups may display distinctive physiologic patterns. Patients who already acquired hyperchloremia or with missing chloride data on day 1 were further excluded. The resulting 33,330 unique ICU stays were split over the 70:30 ratio to form the original training and testing cohorts. 
\subsection*{Performance Comparison and Evaluation}
To test the predictive power of SDOH features, we constructed three feature combinations—the full set (34 features), SDOH Only set (4 features), and Labs Only set (30 features)—with the same original training cohort. To validate and compare model performance among subgroups, we divided the original testing cohort into ten test sets based on the SDOH traits (4 races, 5 insurance types, and 2 genders).  Overall, 12 models were constructed, and 132 sets of testing statistics were obtained. For reporting, we only included the 44 model-subgroup pairs derived from the 4 classifiers trained with the full features sets. Classifier performance was measured with precision, recall, F1-scores, and ROC-AUC. The two main test comparison outcomes are: 1. comparisons among three sets of models built with the full feature set, SDOH feature set, and Labs feature set; and 2.  comparisons between the SDOH subgroup and the baseline full test set metrics. To mitigate the limitation of the subgroup test sets' relatively small size, we derived testing statistics from bootstrapping each subgroup, running the test 1,000 times, and computing the mean of the resulting AUC score distribution. The average AUC scores of different model-subgroup combinations were then compared between the full test set (baseline) and subgroups. Statistical significance (p-value) of difference in model performance was computed using Permutation Test. 

To test the hypothesis of whether using more relevant data in the training process will result in better testing performance in the underrepresented subgroups, we built and tested 40 additionally models (10 SDOH groups, 4 classifiers each) on single-SDOH feature training and testing sets. Specifically, we selected data points in one SDOH domain (for instance, an all-Black-patient cohort) from the original testing set for each SDOH features to train a domain-specific model and tested it on the same SDOH test set derived from the full testing cohort by similar means. The rationale behind this approach is to test if the Black-patient-cohort-trained model would yield improved performance compared to the general model trained on all patients—when testing on the same black patients only test set. For this task, training and testing AUCs were reported for each model-SDOH pair. The Insurance-Self Pay feature was excluded from the model building due to its small sample size. 
\clearpage

\subsection*{Feature Analysis and Visualizations}
To further interpret classifier performance, we performed secondary analyses on the feature importance of our best performing XGBoost model using the SHAP (SHapley Additive explanations) method—a game-theoretic approach that explains individual predictions made by complex models.\cite{ref13} In addition to assessing the full model, we built and performed SHAP analysis on 4 additional classifiers trained with single-race data—to understand if certain feature plays a more important role in the racially distinctive populations. The minority-race-trained models allow us insights into the characteristics of underrepresented populations and potential explanations for the varied between-group performance displayed by the full model. Additional figures were created to investigate the distribution of chloride level measured after the first 24 hours of ICU admission in SDOH subgroups. 

\section*{RESULTS}
\subsection*{Study Population}
The study population consisted of 333,30 unique ICU stays. The self-reported age, race, gender, and insurance status were shown in Table 1. Notice that self-reported race was missing in 4,269 (12.8\%) encounters. Overall, the original cohort had a case rate of 56.0\%, an average patient age of 65.5, and was predominantly male (57.2\%). The demographic features were relatively constant across the four race subgroups. However, we observed that the Hispanic population was noticeably younger (52.8) and had fewer female patients (36.4\%), while Asian patients have the highest case rate of 9.2\%.  About half of the stays were covered by Medicare, followed by private insurance payments (31.6\%).   
\begin{table}[]
\centering
\begin{tabular}{llllll}
\hline
                          & \textit{\textbf{Black}} & \textit{\textbf{Asian}} & \textit{\textbf{Hispanic}} & \textit{\textbf{White}} & \textit{\textbf{Total*}} \\ \hline
N                         & 3283                    & 688                     & 1180                       & 23910                   & 33330                    \\
Female, n(\%)             & 1792 (54.6)             & 285 (41.4)              & 430 (36.4)                 & 10109 (42.3)            & 14277 (42.8)             \\
Age, median (IQR)         & 59.4 (23.9)             & 64.4 (26.5)             & 52.8 (24.1)                & 66.9 (24.6)             & 65.5 (25.1)              \\
Hyperchloremia, n(\%)     & 175 (5.3)               & 63 (9.2)                & 80 (6.8)                   & 1382 (5.8)              & 1992 (6.0)               \\
\multicolumn{6}{l}{\textbf{Insurance}}                                                                                                                          \\
Government-Insured, n(\%) & 122 (3.7)               & 30 (4.4)                & 99 (8.4)                   & 525 (2.2)               & 917 (2.8)                \\
Medicare, n(\%)           & 1790 (54.5)             & 346 (50.3)              & 505 (42.8)                 & 13612 (56.9)            & 18559 (55.7)             \\
Medicaid, n(\%)           & 488 (14.9)              & 119 (17.3)              & 223 (18.9)                 & 1770 (7.4)              & 2958 (8.8)               \\
Private, n(\%)            & 848 (25.8)              & 189 (27.5)              & 321 (27.2)                 & 7793 (32.6)             & 10524 (31.6)             \\
Self-Pay, n(\%)           & 35 (1.1)                & 4 (0.6)                 & 32 (2.7)                   & 209 (0.9)               & 371 (1.1)                \\ \hline
\end{tabular}
\caption{Patient Demographics by Races }
\label{table 1}
\end{table}
\subsection*{Model Comparison}
A total of 12 models (4 classifiers, each trained on 3 sets of features) were developed and tested on the full test set, table 2. We compared the mean AUC scores derived from iteratively testing each model 1,000 times with bootstrapped test sample. Two sets of comparisons were conducted: one between the Full-Feature set trained models and the SDOH-Only trained ones and another between the Full Feature models and Lab Only ones. The performance metrics were compared between each model and their respective baseline Full Feature Model; a p-value was computed from the permutation test.  

Using SDOH Only features alone yielded significantly lower AUCs in the 0.55-0.6 range comparing to the other two groups. Additionally, the results revealed that all Labs Only models performed significantly worse compared to the Full-feature ones except for Multilayer Perception (p=0.18), although the differences in AUCs were small. This implies that adding the SDOH features onto traditional clinical ones improves the prediction outcomes of the classifiers. Overall, the XGBoost classifier trained on the Full Feature set achieved the highest test performance (AUC=0.7977), while Multilayer Perception performed consistently worse across all three feature groups. 

\begin{table}[h]
\centering
\begin{tabular}{@{}lllllllll@{}}
\toprule
                          & \multicolumn{2}{l}{\textit{\textbf{Full Features Set}}} & \multicolumn{2}{l}{\textit{\textbf{Health Determ. Set}}} & \textit{\textbf{p value 1}} & \multicolumn{2}{l}{\textit{\textbf{Labs Only}}} & \textit{\textbf{P value 2}} \\ \midrule
\textit{\textbf{Model}}   & \textit{Train}              & \textit{Test}             & \textit{Train}              & \textit{Test}              & \textit{}                  & \textit{Train}          & \textit{Test}         & \textit{}                  \\
\textit{Ridge Regression} & 0.7945                      & 0.7861                    & 0.5596                      & 0.5498                     & \textless{}0.001           & 0.7842                  & 0.7754                & \textless{}0.001           \\
\textit{Random Forest}    & 0.9061                      & 0.7955                    & 0.7844                      & 0.5271                     & \textless{}0.001           & 0.8781                  & 0.7913                & \textless{}0.001           \\
\textit{MLP}              & 0.7859                      & 0.7831                    & 0.5857                      & 0.5560                     & \textless{}0.001           & 0.8228                  & 0.7826                & 0.181          \\
\textit{XGBoost}          & 0.8466                      & 0.7977                    & 0.6243                      & 0.5572                     & \textless{}0.001           & 0.8300                  & 0.7931                & \textless{}0.001           \\ \bottomrule
\end{tabular}
\caption{Performance Comparison among classifiers trained with three features sets. }
\label{table 2}
\end{table}
\subsection*{Subgroup Performance}
We subsequently tested the four Full Feature models’ performance in the 11 demographic test sets. A total of 44 sets of subgroup-model metrics were reported and the respective AUC scores were reported in Table 3. We computed the p-values to assess whether the subgroups’ AUCs were significantly different than the model performance on the full testing set. We observed that 4 out of the 44 subgroup-model pairs (RF-White, RF-Medicare, XGB-White, and XGB-female) achieved similar performance as the baseline full testing test. In other words, all four classifiers performed differently in most of the demographic subgroups. Overall, the XGBoost trained on the full feature set performed the best (AUC= 0.7978) while the MLP trained on SDOH only achieved the least (AUC = 0.5743). Additionally, the SDOH only models perform significantly less well than Labs only one. 

\begin{table}[ht]
\centering
\begin{tabular}{@{}lcccccccc@{}}
\toprule
             & \multicolumn{2}{c}{\textbf{LR}}                       & \multicolumn{2}{c}{\textbf{MLP}}                      & \multicolumn{2}{c}{\textbf{RF}}                       & \multicolumn{2}{c}{\textbf{XGBoost}}                  \\ \midrule
             & \textit{\textbf{ROC-AUC}} & \textit{\textbf{P-value}} & \textit{\textbf{ROC-AUC}} & \textit{\textbf{p-value}} & \textit{\textbf{ROC-AUC}} & \textit{\textbf{p-value}} & \textit{\textbf{ROC-AUC}} & \textit{\textbf{p-value}} \\
Training     & 0.7847                    &                           & 0.7993                    &                           & 0.9441                    &                           & 0.8466                    &                           \\
Full Testing & 0.7781                    &                           & 0.7844                    &                           & 0.7936                    &                           & 0.7975                    &                           \\
\multicolumn{9}{l}{\textbf{Race}}                                                                                                                                                                                                            \\
Black        & 0.8354                    & \textless{}0.001          & 0.8359                    & \textless{}0.001          & 0.8278                    & \textless{}0.001          & 0.8107                    & \textless{}0.001          \\
Asian        & 0.8328                    & \textless{}0.001          & 0.8202                    & \textless{}0.001          & 0.8683                    & \textless{}0.001          & 0.8544                    & \textless{}0.001          \\
Hispanic     & 0.7483                    & \textless{}0.001          & 0.7636                    & \textless{}0.001          & 0.7675                    & \textless{}0.001          & 0.7930                    & \textless{}0.001          \\
White        & 0.7759                    & \textless{}0.001          & 0.7820                    & \textless{}0.001          & 0.7934                    & 0.6290                    & 0.7968                    & 0.0970                    \\
\multicolumn{9}{l}{\textbf{Gender}}                                                                                                                                                                                                          \\
Female       & 0.7729                    & \textless{}0.001          & 0.7782                    & \textless{}0.001          & 0.7924                    & 0.0100                    & 0.7980                    & 0.8410                    \\
Male         & 0.7845                    & \textless{}0.001          & 0.7853                    & 0.0290                    & 0.7915                    & \textless{}0.001          & 0.7943                    & \textless{}0.001          \\
\multicolumn{9}{l}{\textbf{Insurance}}                                                                                                                                                                                                       \\
Government   & 0.8510                    & \textless{}0.001          & 0.8715                    & \textless{}0.001          & 0.8738                    & \textless{}0.001          & 0.8993                    & \textless{}0.001          \\
Medicare     & 0.7716                    & \textless{}0.001          & 0.7784                    & \textless{}0.001          & 0.7936                    & 0.8380                    & 0.7991                    & 0.0040                    \\
Medicaid     & 0.8001                    & \textless{}0.001          & 0.7873                    & 0.0100                    & 0.7861                    & \textless{}0.001          & 0.7882                    & \textless{}0.001          \\
Private      & 0.7751                    & \textless{}0.001          & 0.7824                    & \textless{}0.001          & 0.7877                    & \textless{}0.001          & 0.7847                    & \textless{}0.001          \\
Self-Pay     & 0.8417                    & \textless{}0.001          & 0.8500                    & \textless{}0.001          & 0.7806                    & \textless{}0.001          & 0.7250                    & \textless{}0.001          \\ \bottomrule
\end{tabular}
\caption{Performance comparison between the SDOH subgroups and the full testing set AUCs. }
\label{table 3}
\end{table}
\begin{figure}[]
\centering
\includegraphics[width=\textwidth,height=\textheight,keepaspectratio]{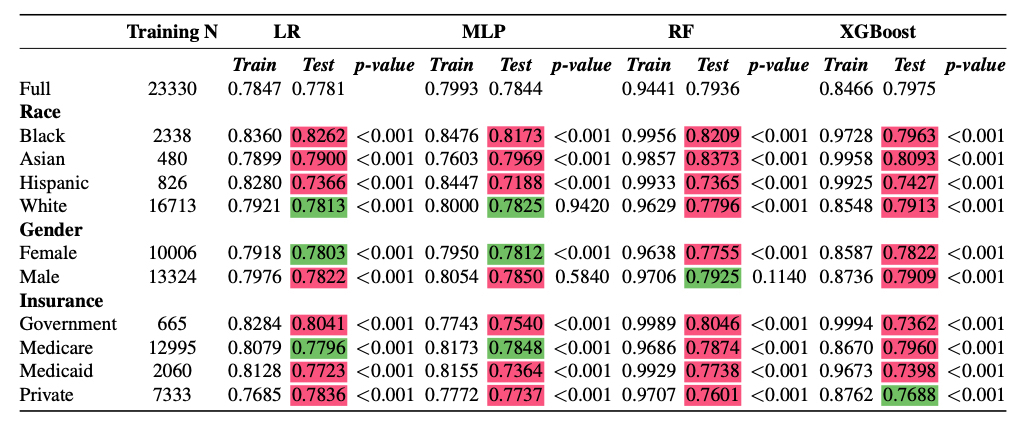}
\caption{Training and Testing AUCs of the SDOH-specific models. }
\label{table 4}
\end{figure}

A closer look at the race subgroups’ performance revealed that all four models yielded significantly lower AUC scores when testing on the Hispanic population, as compared to other racial groups. Conversely, the models performed constantly better on Black and Asian populations: in all four models, the AUC scores of the two race groups were all in the above 0.80 AUC range, which were even higher than the performance on the training and full testing set performances. The White-patient-only group, which made up over 70\% of the original testing cohort, achieved highly similar results as the full test set metrics. In terms of gender, AUCs were similar (~0.77-79) across all four models in both genders. No consistent patterns were observed: the LR and MLP models achieved slightly higher AUC scores in male patients when compared to both the full and female subgroups; however, the advantage was not seen in the RF and XGBoost models.  The Insurance subgroups displayed a more varied AUC distribution. All four classifiers achieved the highest performance with government-insured patients and the lowest in the privately insured subgroup. Performance on Medicare and Medicaid patients was less likely to differ from the baseline. XGBoost was the least efficient model when predicting insurance-specific groups (4 out of 5 groups under-performed comparing to baseline). 

As an additional step to validate whether domain-specific models (i.e., trained and tested on the same SDOH-only sets) outperform the full-patients model, 40 SDOH classifiers were built and tested. The resulting training and testing AUCs were reported in figure 2. Of the all SDOH-specific models, only 8 marginally outperformed the full-patient model while 32 obtains lower AUC scores. Significance test between each SDOH-specific model and the full patient model’s performance on the same SDOH test shows that the performance disparity was only insignificant in three cases (MLP-male, MLP-White, and RF-Male).  We observed no clear pattern in the outcome improvements with respect to classifier or a specific patient population.  This finding indicated that having SDOH-specific models would not necessarily improve testing performances in the respective minority population in this cohort of study. 
\subsection*{Secondary Analysis}
We applied the SHAP method on our best-performing XGBoost classifier and rendered the SHAP summary plot, which offered a bird-eye view of feature importance from high to low and allowed a straightforward interpretation of which direction did a feature push the prediction. The color represented the value of each feature, and the horizontal location indicated the magnitude of the SHAP value (i.e., whether the effect of that value caused a higher or lower prediction). In the simplest term, chloride level measured on day one of their ICU stay was the most utilized feature; patients with high chloride measurement on day one was more likely to develop hyperchloremia in the following days, which agreed with common-sense understanding of the disease. We also observed that receiving ventilation treatment and a greater value of the highest daily chloride input were linked to a higher chance of developing Hyperchloremia. In terms of age, the classifier predicted older patients as hyperchloremic more often. The plot revealed that our XGBoost classifier depended heavily on a handful of features while essentially omitting the others.

\begin{figure}[h]
\centering
\includegraphics[width=350pt,height=191pt]{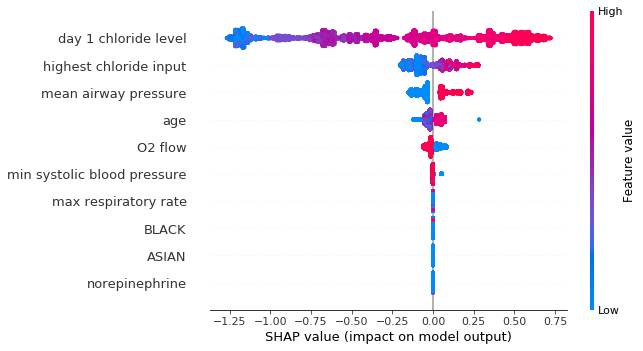}
\caption{SHAP Summary Plot for full models}
\label{fig2}
\end{figure}
\begin{figure}[h]
\centering
\includegraphics[width=\textwidth,height=\textheight,keepaspectratio]{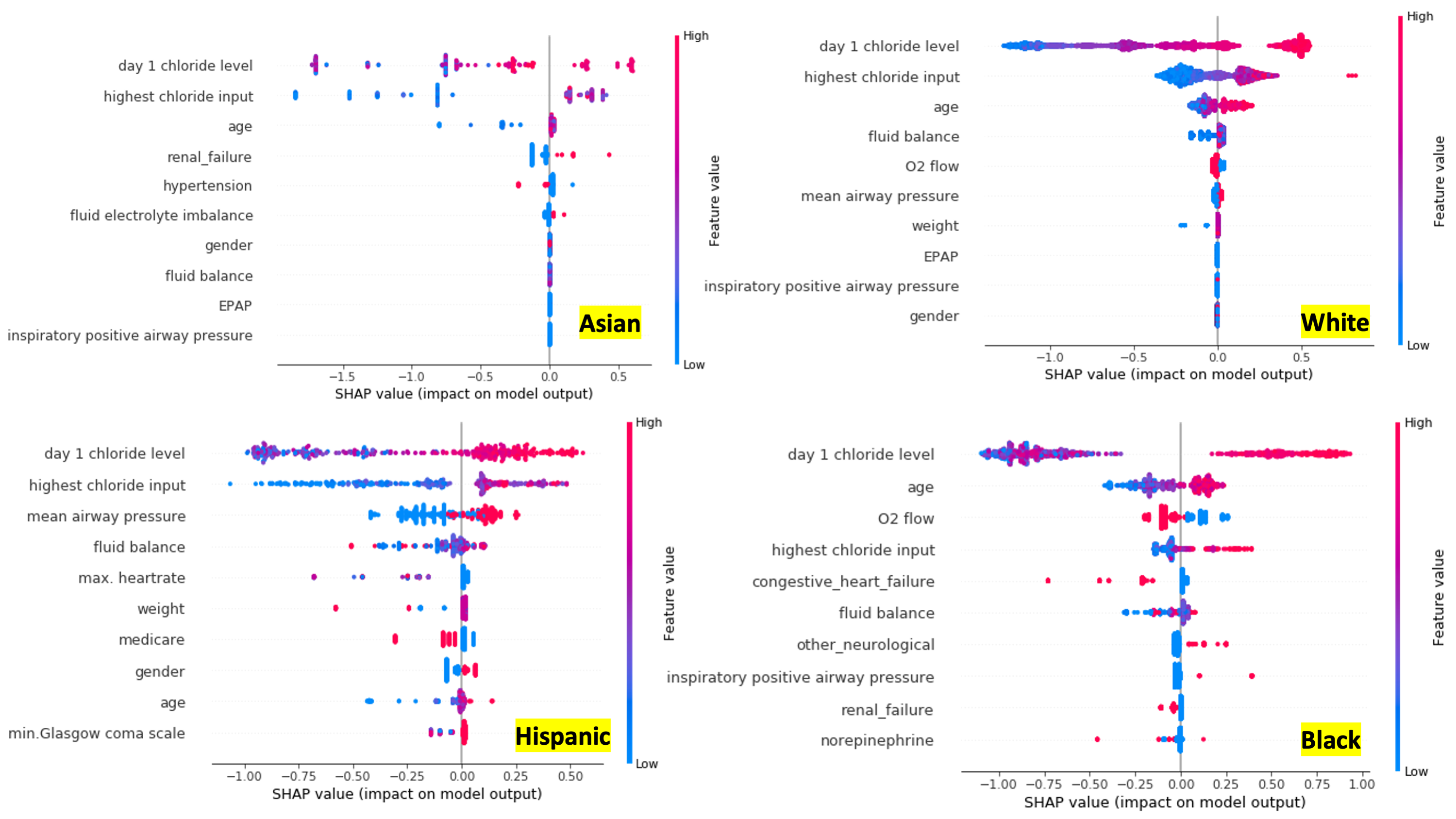}
\caption{SHAP Summary Plots for racial-group trained models}
\label{fig3}
\end{figure}
In the subsequent analysis of four single-race trained XGBoost classifiers (Black, Asian, Hispanic, White), chloride level on day one, the highest chloride input, and ventilation methods heavily contributed to the prediction outcome in all four race-based models.  Age ranked among the top predictors in all race-only models; however, it played a less important role in the Hispanic model than in  other racial models. Gender was critical in the prediction task in all except the Black model. Unlike the full model, where lab results played a central role in prediction, the minority-race-trained classifiers relied more heavily on demographic and co-morbidity upon admission. For instance, co-morbidities were among the top predictors in Asian and Black models. In the Asian-trained model, having renal failure as a co-morbidity on admission indicated a greater risk of developing hyperchloremia while having hypertension as a co-morbidity on admission was linked to a decrease in the predicted likelihood. In the Black model, congestive heart failure, renal failure, and neurological conditions were shown to have a significant impact on prediction outcome as well. A further look at the unique predictors in each race model revealed that the Hispanic model had the most group-specific top features (Medicare, Maximum heart rate, and Minimum Glasgow Coma Scale), while the White model's top features appear in at least one other model. 
\section*{DISCUSSION}
\subsection*{Main findings}
In this study, we tested the predictive power of social determinants of health indicators and observed that adding the features significantly improved the performance of models trained with clinical indicators alone. Through extensive subgroup testing, we identified widespread performance variation in each classifier among different gender, race, and insurance status subgroups. We found a general trend that all subgroup classifiers' AUC scores on the White test population tend to be highly similar to the AUCs trained from all patients. The classifiers constantly under-performed on Hispanic patients. The performance in the Black and Asian subgroups, on the other hand, were higher than the full testing set metrics in the majority of cases. Additionally, we observed that SDOH-specific model does not guarantee improved performance in a given subgroup. This finding reminded us the trade-off between the specificity of data and sample size. The benefits of having more relevant training data in the SDOH-trained models could be offset by the smaller sample size used in model training. The feature importance analysis reveals that certain intervention strategies such as ventilation, age, and chloride measurements on the first day of ICU admission are highly associated with the chance of developing hyperchloremia, although this trend might not imply a causal relationship. 

Hyperchloremic is a highly prevalent condition observed in the ICU, afflicting approximately a third of the population as reported by several studies.\cite{ref10,ref13} However, despite the increased interest in this area over the past few years, there has been a lack of understanding of the disease with respect to the social determinants of the health dimension. We found that the Asian population in our dataset had the highest percentage of Hyperchloremia patients (9.16\%) that almost doubled the population average (5.98\%), while the Black ranked the lowest with a prevalence rate of 5.53\%.  In the SHAP analysis, we observed variations in the ranking and weights of the predictors’ feature importance in the four single-race training populations. In other words, the classifier utilized different combinations of features when making predictions in a minority population, which could imply that different racial groups possess distinctive disease profiles. 

Previous studies presented mixed evidence on the presence and severity of machine learning biases in different areas of healthcare applications. We summarized the possible reasons that could lead to the divide in findings as follows. First, there exists only a handful of studies that tested the models with demographic subgroups, and each study originated from different areas of clinical research. We were unable to identify two or more comparable studies in the same realm, thus preventing us from corroborating the research findings. Second, we observed a lack of consensus on how evaluation tasks should be designed and what methods to use in measuring the extent of biases. For instance, we performed significance testing to determine whether the subgroup performances were statistically different from the baseline values. In other studies, however, conclusions were drawn by eyeballing the absolute values of different performance metrics without further validation. To address these gaps, a more rigorous evaluation framework needs to be in place. 
\subsection*{Strengths and limitations}
This study is the first to assess the performance bias in machine learning classifiers designed to predict Hyperchloremia in ICU populations. However, our assessment alone could not tell a full story of the potential bias in clinical machine learning applications. A more comprehensive view could only be obtained when more researchers would incorporate similar model testing strategies in their studies. 

While consistent trends in performance disparities were observed in subgroup testing, the evidence might not be rigorous enough for us to draw any definitive conclusion without conducting tests on external validation datasets. The trends might as well be anecdotal, given that the relatively small sample size of the racial minorities in the study cohort may not represent the entire under-represented populations. Due to the limited amount of demographic information that comes with the MIMIC III dataset, we used insurance status as a sole proxy to SES status. We believe that a more comprehensive definition of SES, as well as the availability of a wider choice of SDOH features (educational status, urbanity, health literacy, etc.), would improve the validity of our statement. Nevertheless, we made a tentative step into dissecting performance biases in this study and flagged out areas that are in desperate need of further investigations. Additionally, we did not explore possible options to proactively address the foreseeable biases (i.e., datasets with unbalanced racial composition) in the model building process due to the limited scope of the paper.
\subsection*{Future Directions}
Our study was designed to test the equal performance statement, and indeed identified between-subgroup variations in prediction accuracy. However, the actual impact of the biased performance on patient outcomes or medical resource allocation could not be inferred from these analyses alone. In other words, we would not be able to conclude whether the less inferior model performance on the minority samples directly contributed to unfavorable outcomes in the respective population. Thus, one fruitful direction is to quantify the actual impact of machine learning performance and to prove the connection between biases and measurable clinical outcomes. In addition to addressing biases in algorithm design and implementation, another promising area of research is to enhance interpretability and transparency in machine learning. An overwhelming number of widely-used models still remain as uninterpretable ‘algorithmic black boxes’, which offered little relevant knowledge to the users.  In this paper, we made attempts to interpret model performance with respect to demographic features, such as race and gender, using SHAP analysis. However, the exploratory nature of the aforementioned methods forbids us from drawing a conclusion on the relationship between model performance and demographic profiles. Thus, we would urge future researchers to focus on interpreting machine learning models’ performance in different subgroups. 
\section*{Conclusion}
The widespread adoption of Machine Learning in medicine has far-reaching implications on healthcare outcomes. By sharing the lessons learned from this preliminary assessment, we hope to bring the highly relevant yet oft-neglected problem of biases in Machine Learning into the current discussion around the fair and responsible use of technology in healthcare. Extensive testing on demographically diverse subpopulation offers richer insights on the applicability of an ML model, thus helping clinical practitioners to better evaluate the quality and applicability of a proposed model. We believe that subgroup reporting is a promising strategy that would guide the Safe and responsible implementations of algorithm-based decision-support tools. 
\makeatletter

\renewcommand{\@biblabel}[1]{\hfill #1.}
\makeatother
\bibliographystyle{unsrt}

\begin{thebibliography}{1}
\setlength\itemsep{-0.1em}

\bibitem{ref1a}
Wang, H., Li, Y., Ning, H., Wilkins, J., Lloyd-Jones, D. and Luo, Y., Using Machine Learning to Integrate Socio-Behavioral Factors in Predicting Cardiovascular-Related Mortality Risk. In MedInfo 2019, August. (pp. 433-437).

\bibitem{ref1}
Rajkomar A, Hardt M, Howell MD, Corrado G, Chin MH. Ensuring fairness in machine learning to advance health equity. Annals of internal medicine. 2018 Dec 18;169(12):866-72.

\bibitem{ref2}
Gianfrancesco MA, Tamang S, Yazdany J, Schmajuk G. Potential biases in machine learning algorithms using electronic health record data. JAMA internal medicine. 2018 Nov 1;178(11):1544-7.

\bibitem{ref3}
Buolamwini J, Gebru T. Gender shades: Intersectional accuracy disparities in commercial gender classification. In Conference on fairness, accountability and transparency 2018 Jan 21 (pp. 77-91). PMLR.

\bibitem{ref4}
Noseworthy PA, Attia ZI, Brewer LC, Hayes SN, Yao X, Kapa S, Friedman PA, Lopez-Jimenez F. Assessing and mitigating bias in medical artificial intelligence: the effects of race and ethnicity on a deep learning model for ECG analysis. Circulation: Arrhythmia and Electrophysiology. 2020 Mar;13(3):e007988.

\bibitem{ref5}
Chen IY, Szolovits P, Ghassemi M. Can AI help reduce disparities in general medical and mental health care?. AMA journal of ethics. 2019 Feb 1;21(2):167-79.

\bibitem{ref6}
Obermeyer Z, Powers B, Vogeli C, Mullainathan S. Dissecting racial bias in an algorithm used to manage the health of populations. Science. 2019 Oct 25;366(6464):447-53.

\bibitem{ref6a}
Bhavani SV, Luo Y, Miller WD, Sanchez-Pinto LN, Han X, Mao C, Sandikci B, Peek ME, Coopersmith CM, Michelson KN, Parker WF. Simulation of ventilator allocation in critically ill patients with COVID-19. American Journal of Respiratory and Critical Care Medicine. 2021 204(10) (pp. 1224-1227).

\bibitem{ref7}
Reuter DA, Chappell D, Perel A. The dark sides of fluid administration in the critically ill patient. Intensive care medicine. 2018 Jul;44(7):1138-40.

\bibitem{ref8}
Sen A, Keener CM, Sileanu FE, Foldes E, Clermont G, Murugan R, Kellum JA. Chloride content of fluids used for large volume resuscitation is associated with reduced survival. Critical care medicine. 2017 Feb;45(2):e146.

\bibitem{ref8a}
Li Y, Yao L, Mao C, Srivastava A, Jiang X, Luo Y. Early prediction of acute kidney injury in critical care setting using clinical notes. In2018 IEEE International Conference on Bioinformatics and Biomedicine (BIBM) 2018 Dec 3 (pp. 683-686).

\bibitem{ref9}
Suetrong B, Pisitsak C, Boyd JH, Russell JA, Walley KR. Hyperchloremia and moderate increase in serum chloride are associated with acute kidney injury in severe sepsis and septic shock patients. Critical Care. 2016 Dec;20(1):1-8.

\bibitem{ref10}
Neyra JA, Canepa-Escaro F, Li X, Manllo J, Adams-Huet B, Yee J, Yessayan L. Association of hyperchloremia with hospital mortality in critically ill septic patients. Critical care medicine. 2015 Sep;43(9):1938.

\bibitem{ref11}
Jolley SE, Hough CL, Clermont G, Hayden D, Hou S, Schoenfeld D, Smith NL, Thompson BT, Bernard GR, Angus DC. Relationship between race and the effect of fluids on long-term mortality after acute respiratory distress syndrome. Secondary Analysis of the National Heart, Lung, and Blood Institute Fluid and Catheter Treatment Trial. Annals of the American Thoracic Society. 2017 Sep;14(9):1443-9.

\bibitem{ref12}
Yeh P, Pan Y, Sanchez-Pinto LN, Luo Y. Hyperchloremia in critically ill patients: association with outcomes and prediction using electronic health record data. BMC Medical Informatics and Decision Making. 2020 Dec;20(14):1-0.

\bibitem{ref13}
Lundberg S, Lee SI. A unified approach to interpreting model predictions. arXiv preprint arXiv:1705.07874. 2017 May 22.

\bibitem{ref14}
Shao M, Li G, Sarvottam K, Wang S, Thongprayoon C, Dong Y, Gajic O, Kashani K. Dyschloremia is a risk factor for the development of acute kidney injury in critically ill patients. PloS one. 2016 Aug 4;11(8):e0160322.



\end{thebibliography}

\end{document}